\documentclass[10pt,twocolumn,letterpaper]{article}
\usepackage[accsupp]{axessibility}  
\usepackage{iccv}
\usepackage{times}
\usepackage{epsfig}
\usepackage{graphicx}
\usepackage{amsmath}
\usepackage{amssymb}

\usepackage{booktabs}
\usepackage{float}
\usepackage{url}
\usepackage{multirow}
\usepackage{array}
\usepackage{color}
\usepackage{bbding}
\usepackage{xspace}
\usepackage{wrapfig}
\usepackage{pifont}
\usepackage{colortbl}
\usepackage{algorithm}
\usepackage{algpseudocode}
\usepackage{indentfirst} 
\usepackage{makecell}
\usepackage{cancel}
\usepackage{listings}
\usepackage{bm}
\usepackage{etoolbox}

\usepackage[pagebackref=true,breaklinks=true,letterpaper=true,colorlinks,bookmarks=false]{hyperref}

\iccvfinalcopy 
\newlength\savewidth\newcommand\shline{\noalign{\global\savewidth\arrayrulewidth
  \global\arrayrulewidth 1pt}\hline\noalign{\global\arrayrulewidth\savewidth}}
\newcommand{\tablestyle}[2]{\setlength{\tabcolsep}{#1}\renewcommand{\arraystretch}{#2}\centering\footnotesize}

\newcolumntype{x}[1]{>{\centering\arraybackslash}p{#1pt}}
\newcolumntype{y}[1]{>{\raggedright\arraybackslash}p{#1pt}}
\newcolumntype{z}[1]{>{\raggedleft\arraybackslash}p{#1pt}}

\def\ourmethod{MixVoxels\xspace}
\def\codedescrip{Codes and trained models are available at \url{https://github.com/fengres/mixvoxels}.\xspace}
\usepackage[capitalize]{cleveref}
\crefname{section}{Sec.}{Secs.}
\Crefname{section}{Section}{Sections}
\Crefname{table}{Table}{Tables}
\crefname{table}{Tab.}{Tabs.}

\def\iccvPaperID{5238} 

\ificcvfinal\fi

\begin{document}

\title{Mixed Neural Voxels for Fast Multi-view Video Synthesis}

\author{
        Feng Wang$ ^{1}$ 
        ~~~~~
        Sinan Tan$ ^{1} $ 
        ~~~~~
        Xinghang Li$ ^{1}$
        ~~~~~
        Zeyue Tian$ ^{2} $
        ~~~~~
        Yafei Song$ ^{3} $
        ~~~~~
        Huaping Liu$ ^{1}$ \thanks{Corresponding author.} 
        \vspace{0.5em}
        \\~
        $ ^1 $Beijing National Research Center for Information Science and Technology(BNRist), \\
        Department of Computer Science and Technology, Tsinghua University\\
        ~~~~
        $ ^2 $Hong Kong University of Science and Technology  
        \\
        $ ^3$ XR Lab, DAMO Academy, Alibaba Group
        \\
        \small
        \texttt{wang-f20@mails.tsinghua.edu.cn}, \texttt{hpliu@tsinghua.edu.cn}
}

\maketitle
\ificcvfinal\thispagestyle{empty}\fi

\begin{abstract}
Synthesizing high-fidelity videos from real-world multi-view input is challenging due to the complexities of real-world environments and high-dynamic movements. Previous works based on neural radiance fields have demonstrated high-quality reconstructions of dynamic scenes. However, training such models on real-world scenes is time-consuming, usually taking days or weeks. In this paper, we present a novel method named \ourmethod to efficiently represent dynamic scenes, enabling fast training and rendering speed. The proposed \ourmethod represents the $4$D dynamic scenes as a mixture of static and dynamic voxels and processes them with different networks. In this way, the computation of the required modalities for static voxels can be processed by a lightweight model, which essentially reduces the amount of computation as many daily dynamic scenes are dominated by static backgrounds. To distinguish the two kinds of voxels, we propose a novel variation field to estimate the temporal variance of each voxel. For the dynamic representations, we design an inner product time query method to efficiently query multiple time steps, which is essential to recover the high-dynamic movements. As a result, with 15 minutes of training for dynamic scenes with inputs of 300-frame videos, \ourmethod achieves better PSNR than previous methods. For rendering, \ourmethod can render a novel view video with 1K resolution at 37 fps. \codedescrip
\end{abstract}
\section{Introduction}
\label{sec:intro}
Dynamic scene reconstruction from multi-view videos is a critical and challenging problem, with many potential applications such as interactively free-viewpoint control for movies, cinematic effects like freeze-frame \textit{bullet time}, novel view replays for sporting events, and various potential VR/AR applications. Recently, neural radiance fields \cite{mildenhall2021nerf} have demonstrated the possibility of rendering photo-realistic novel views for static scenes, with physically motivated 3D density and radiance modelling. Many methods \cite{li2022neural,li2021neural, xian2021space, gao2021dynamic, du2021neural, pumarola2021d, park2021nerfies, park2021hypernerf, tretschk2021non} extend the neural radiance fields to dynamic scenes with additional time queries or an explicit deformation field. Many of these methods focus on the monocular input video setting on relatively simple dynamic scenes. To model more complex real-world dynamic scenes, a more practical solution is to use multi-view synchronized videos to provide dense spatial-temporal supervisions \cite{zitnick2004high, lombardi2019neural, bansal20204d, li2022neural}. 

\begin{figure}[t]
\centering
\includegraphics[width=1.0\linewidth]{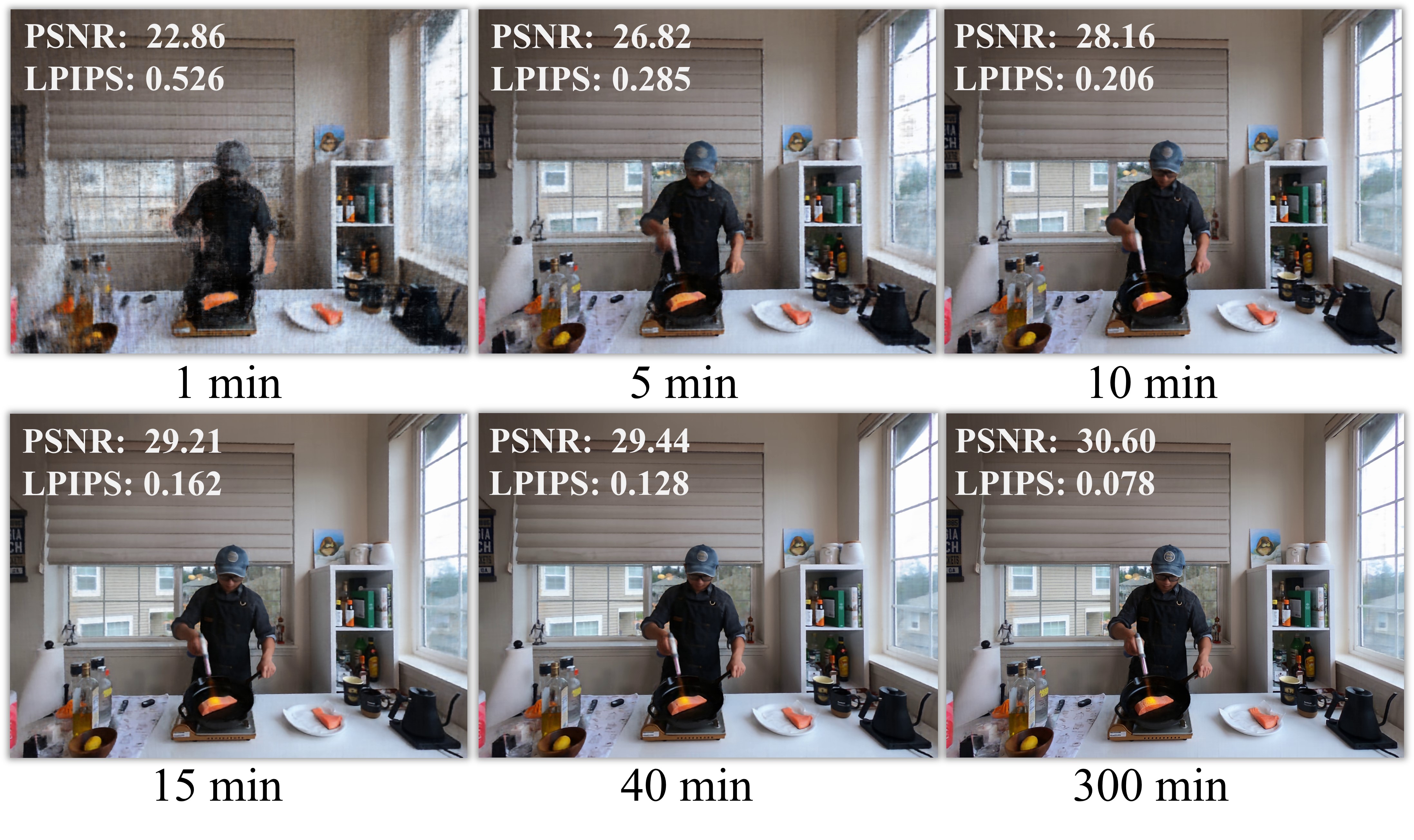}\\
\vspace{-0.4em}
\caption{Our method enables rapid reconstruction of 4D dynamic scenes. We visualize the rendering results with different training schedules. With only 15 minutes of training, our approach achieves comparable PSNRs to other methods. Increasing the training time further enhances the ability to recover fine details.}
\label{fig:teaser}
\vspace{-1.8em}
\end{figure}

Recently, Li et al. \cite{li2022neural} propose a real-world dynamic scene dataset including many challenging situations such as objects of high specularity, topology changes, and volumetric effects. They address the problem by a hierarchical training scheme and the ray importance sampling strategies. Although significant improvements have been achieved, some challenges still exist: (1) The training and rendering take a lot of time and computation resources. (2) Highly dynamic scenes with complex motions are still difficult to track. 

In this paper, we focus on the multi-view 3D video synthesis problem and present a novel method named \ourmethod to address the above two challenges. The proposed \ourmethod is based on the explicit voxel-grid representation, which is recently popular due to its fast training and rendering speed on static scenes \cite{yu2021plenoxels, sun2022direct, chen2022tensorf, muller2022instant}.
We extend the voxel-grid representations to support dynamic scenes and propose an efficient inner product time querying method that can query a large number of time steps simultaneously, which is essential to recover the sharp details for highly-dynamic objects. 
Additionally, we represent dynamic scenes as a mixed static-dynamic voxel-grid representation. Specifically, the 3D spaces are split into static and dynamic voxels by our proposed variation field. The two components are processed by different models to reduce the redundant computations for the static space. Theoretically, once a dynamic scene consists of some static spaces, the training speed will benefit from the proposed mixed voxels. For a variety of events that occur in the physical world, the static components of environments are dominated in most cases, and the mixed voxels will speed up the training significantly in these scenarios. Besides, the separation of voxels makes the time-variant model focus on the dynamic regions, avoiding the time-aware voxels being biased by the static spaces to produce blurred motions. Our empirical validation confirms that the separation enables the model to learn sharp and distinct boundaries in high-dynamic regions. This also frees our method from the complex importance sampling strategies. With these designs, our method is capable of reconstructing a dynamic scene consisting of 300 frames within 15 minutes.
To summarize, the main contributions of this work are:
\begin{itemize}
    \vspace{-0.1em}
    \item We propose a simple yet effective dynamic representation with inner product time querying method that can efficiently query multiple times simultaneously, improving the rendering quality for dynamic objects.
    \vspace{-1.5em}
    \item We design an efficient variation field to separate static and dynamic spaces and present a mixed voxel-grid representation to accelerate training and rendering.
    \vspace{-0.3em}
    \item We conduct qualitative and quantitative experiments to validate our method. As a result, the proposed \ourmethod achieves competitive or better rendering qualities with a $5000 \times$ training speedup compared to implicit dynamic scene representations.
\end{itemize}

\section{Related Works}
\textbf{Novel View Synthesis for Static Scenes.}
Synthesizing novel views for static scenes is a classical and well-studied problem. Different approaches represent the underlying geometric with different representations. Mesh-based methods \cite{buehler2001unstructured, debevec1996modeling, waechter2014let, wood2000surface, riegler2020free, thies2019deferred} represent the scenes with surfaces which is compact and easy to render, while optimizing a mesh to fit complex scenes is challenging. Volume-based methods such as voxel-grid \cite{kutulakos2000theory, seitz1999photorealistic, penner2017soft, lombardi2019neural, sitzmann2019deepvoxels} and multi-plane images (MPIs) \cite{zhou2018stereo, flynn2019deepview, mildenhall2019local, srinivasan2019pushing, srinivasan2020lighthouse, tucker2020single} are more suitable to model the complex and translucent scenes such as smooth and fluid. Particularly, Neural radiance fields \cite{mildenhall2021nerf} represent the scenes with an implicit volumetric neural representation, which employs a coordinate-based neural network to query the density and color for each point. The achieved photo-realistic rendering quality of NeRF led to an explosion of developments in the field. Advances have been made including improving the rendering qualities \cite{tancik2020fourier, barron2021mip}, adapting to more general scenarios \cite{zhang2020nerf++, martin2021nerf, tancik2022block, barron2022mip}, accelerating rendering or training speed \cite{liu2020neural, yu2021plenoctrees, yu2021plenoxels, sun2022direct, chen2022tensorf}, etc.  

\textbf{Novel View Synthesis for Dynamic Scenes.}
Synthesizing novel views for dynamic scenes is a more challenging and applicable problem. Recently, many extensions of NeRF for non-rigid dynamic scenes were proposed, which take a monocular video as input to learn the deformation and radiance fields. These methods can be categorized to modelling deformation implicitly \cite{li2021neural, xian2021space, gao2021dynamic, du2021neural} (learn the non-decoupled deformation and appearance jointly) and explicitly \cite{pumarola2021d, park2021nerfies, park2021hypernerf, tretschk2021non} (learn separated deformation and radiance fields and the deformation fields are usually in the form of relative motion with a canonical static space). 
Though improvements are achieved, reconstructing the complex general scenes is still difficult with only monocular videos. Most methods are constrained to fixed scenes like human-model or restricted motions. For real-world complex scenes, reconstructing from synchronized multi-view videos is more promising due to the dense supervision for every viewpoint and time instant. Earlier works \cite{kanade1997virtualized, zitnick2004high} explore the problem and show the possibility of rendering novel videos from a set of input views. Neural Volumes \cite{lombardi2019neural} proposes to use volumetric representations. They employ an encoder-decoder network to convert input images into a 3D volume, and decode the latent representations by the differentiable ray marching operation. \cite{bansal20204d} presents a data-driven approach for 4D space-time visualization of dynamic scenes by splitting static and dynamic components and using a U-Net structure in screen space to convert intermediate representation to image. Different with this method, our method split the static and dynamic components in the 3D voxel space instead of the pixel space. More recently, DyNerf \cite{li2022neural} uses a temporal-aware neural radiance field to address the problem, and proposes some sampling strategies to train it efficiently. Compared with previous methods, they propose a more complicated real-world dataset and validate their method. For accelerating the reconstruction of dynamic scenes, FourierPlenoctree \cite{wang2022fourier} proposes to model the dynamics in frequency domain, and generate a Plenoctree through multi-view blending to accelerate rendering. They focus on the foreground moving objects extracted via chroma key segmentation, which requires the background should be a pure color (or rely on segmentation algorithms). 
Recently, the acceleration of training and rendering for dynamic scenes has attracted much attention. 
Cocurrent works include StreamRF \cite{li2022streaming} which proposes to accelerate the training of dynamic scenes by modeling the differences of adjacent frames, NeRFPlayer \cite{song2022nerfplayer} which decomposes the dynamic scenes into static, new and deforming components, Hyperreel \cite{attal2023hyperreel} which proposes an efficient sampling network and models keyframes. K-Planes \cite{fridovich2023k} and HexPlanes \cite{cao2023hexplane} decompose the 4D dynamic scenes into different 2D representations.

\textbf{Acceleration of Neural Radiance Fields.} While Neural radiance fields can render novel views with high fidelity, training and rendering require querying a deep MLP millions of times which is computationally intensive. Many recent methods propose to accelerate the training and rendering speed of NeRF. For rendering, Neural Sparse Voxel Fields \cite{liu2020neural} proposes a voxel-grid representation to skip over many empty regions. PlenOctree \cite{yu2021plenoctrees} accelerates the rendering process by pre-tabulating the NeRF into a PlenOctree and using the spherical harmonic representation of radiance. Derf \cite{rebain2021derf} and Kilonerf \cite{reiser2021kilonerf} propose to accelerate the rendering speed by dividing the scenes into multiple areas, and employ multiple small network in each area. AutoInt \cite{lindell2021autoint} proposes to restructure the MLP network to accelerate the computations of ray integrals, which helps accelerate the rendering speed.
For accelerating the training of NeRF, some methods use explicit voxel-grid representations \cite{yu2021plenoxels, sun2022direct} to accelerate the training process and convergence speed. Instant-NGP \cite{muller2022instant} proposes a multi-resolution hash table structure to accelerate the training. The model sizes of most fast training methods are relatively large due to a large number of voxels. TensoRF \cite{chen2022tensorf} proposes to reduce the model size by factorizing the 4D scene tensor into multiple compact low-rank tensor components.


\section{Method}
\label{sec:approach}
In this section, we introduce the proposed \ourmethod, which represents the 4D dynamic scenes as mixtures of static and dynamic voxels. \cref{fig:arch} illustrates the overview of our method. In the following subsections, we will first introduce the voxel-grid representations for static scenes and our extension to dynamic scenes. Then we introduce the variation field for identifying the dynamic voxels. At last, we introduce the training of \ourmethod.
\subsection{Static Voxel-grid Representation}
\label{sec: static_model}
Neural radiance fields \cite{mildenhall2021nerf} have demonstrated photo-realistic novel viewpoint synthesis, while the training of NeRF requires extensive computation due to millions of neural network queries. For accelerating NeRF, many recent works \cite{liu2020neural, yu2021plenoxels, sun2022direct, chen2022tensorf} have explored the explicit volumetric representation, which avoids the huge amount of computation of querying neural network. Specifically, a 3D scene is split into $N_x \times N_y \times N_z$ voxels. The densities and color features are stored in these voxels and denoted as $\mathcal{S}^{\sigma} \in \mathbb{R}^{N_x \times N_y \times N_z}$ and $\mathcal{S}^{c} \in \mathbb{R}^{N_x \times N_y \times N_z \times C}$. $\mathcal{S}^{\sigma}_{i,j,k}$ and $\mathcal{S}^{c}_{i,j,k}$ represent the learnable density and color feature of the voxel corner at a discrete position $(i, j, k)$. For a continuous position $(x, y, z)$, the representation $\mathcal{S}_{x,y,z}$ can be calculated by interpolating the nearest $8$ discrete positions. A small MLP network $\mathcal{C}_\theta$ is used to parse the color features into RGB values, taking $S^{c}$ and view direction $\bm{d}$ as input. Formally, the density $\sigma$ and color $c$ is formulated as
\begin{equation}
    \sigma(x, y, z) = \mathcal{S}^{\sigma}_{x, y, z}, \quad
    c(x,y,z, \bm{d}) = \mathcal{C}_{\theta}(\mathcal{S}^{c}_{x, y, z}, \bm{d}) .
\end{equation}

\subsection{Dynamic Voxel-grid Representation}
\label{sec: dy_model}
For dynamic scenes, a direct extension is to add the time dimension to the static voxel-grid representation $\mathcal{S}$ explicitly. However, this direct extension is almost memory-prohibitive due to the large and linearly increasing memory footprint. For a 300-frame video, the learned models will occupy $30$ GB of memory and be difficult to train with GPUs due to the limitation of GPU memory.  To address this problem, we propose a spatially explicit and temporally implicit representation to reduce the memory footprint. 
Specifically, we represent the dynamic scene as a 4D learnable voxel-grid $\mathcal{G}^{\sigma} \in \mathbb{R}^{N_x \times N_y \times N_z \times C_1} $ and $\mathcal{G}^{c} \in \mathbb{R}^{N_x \times N_y \times N_z \times C_2}$. Different from the static scene representations, the densities and colors for all time steps are implicitly encoded as compact features stored in each voxel corner. The compact features will be processed by a time-aware projection to acquire density and color for each time step.
Concretely, for the compact density feature $\mathcal{G}^{\sigma}_{x, y, z}$ and color feature $\mathcal{G}^{c}_{x, y, z}$ in any position$(x,y,z)$, we employ two MLPs $\mathcal{T}^{\sigma}_{\theta_1}$ and $\mathcal{T}^c_{\theta_2}$ to increase the feature dimensions for better parsing time-variant density and color. The MLPs here can be viewed as decompressors that decompress the compact low-dimensional voxel-grid features into more tractable ones. Compared with directly storing high-dimensional features in each voxel, the temporally implicit representation reduces the memory footprint significantly since the shared MLPs only increase memory slightly.
\begin{figure}[t]
\centering
\includegraphics[width=1.0\linewidth]{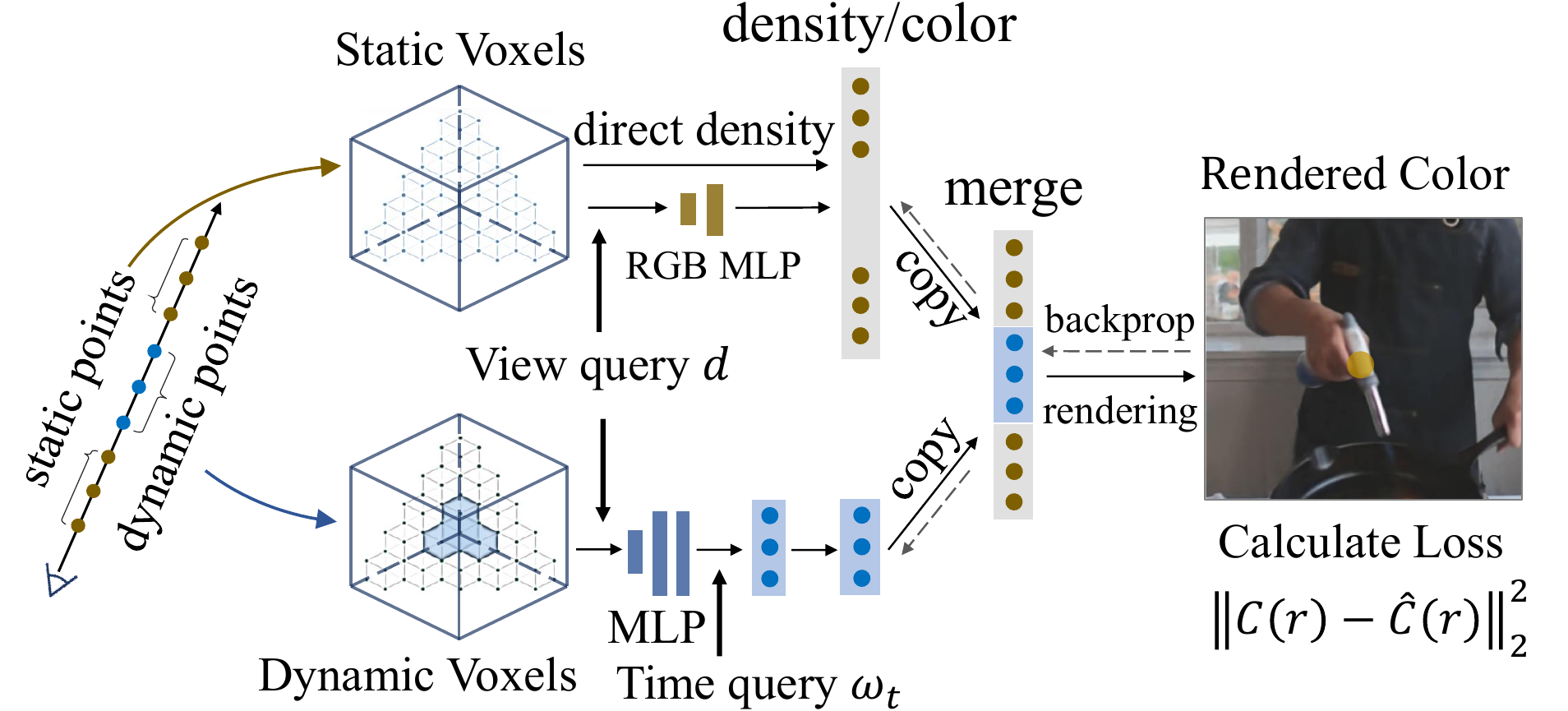}\\
\caption{Overview of our method. Given a ray, we first sample points, and split them into static and dynamic ones using the variation field. After that, we feed these points to the corresponding branches and query the required properties. Then we merge the static output and dynamic output for rendering the ray color. An L2 loss is employed to calculate loss and back-propagate. }
\label{fig:arch}
\vspace{-1.6em}
\end{figure}

\textbf{Inner product time query.} For a discrete time step $t$, we use a learnable time-variant latent representation $\omega_t$ to represent the time query. Instead of concatenating the time query with the intermediate features, we propose to calculate the inner product between the learned time query and the decompressed features as the required output $\sigma$ and $c$. Formally, the density and color of a space-time query $(x,y,z,t)$ are formulated as
\begin{equation}
    \sigma(x, y, z, t) = \omega_{t}^{\sigma} \cdot  \mathcal{T}^{\sigma}_{\theta_1}(\mathcal{G}^{\sigma}_{x,y,z}),
\end{equation}
\begin{equation}
    c(x, y, z, \bm{d}, t) = \omega_{t}^{c} \cdot  \mathcal{T}^{c}_{\theta_2}(\mathcal{G}^{c}_{x,y,z}, \bm{d}).
\end{equation}
In practice, simultaneously querying multiple time steps helps reconstruct the detail of high-dynamic motions and reduces the training iterations to traverse through all time steps. The inner product based query will facilitate the training speed when simultaneously querying many time steps in a training iteration. 
Specifically, we denote the FLOPs of the MLP $\mathcal{T}_{\theta_1}$ and the inner product operation as FLOP$_{mlp}$ and FLOP$_{inn}$, respectively. For a $T$-frame video, the FLOPs of the concatenation query \cite{li2022neural} is larger than $T \cdot$ FLOP$_{mlp}$ (due to the extra temporal embedding dimension), while the FLOPs of our inner product query is only FLOP$_{mlp}$ + $T \cdot$ FLOP$_{inn}$ (FLOP$_{mlp} >> $ FLOP$_{inn}$).

\subsection{Variation Field}
\label{subsec:variance_field}
In this subsection, we introduce the variation field to identify which voxels in the 3D space are dynamic, \textit{i.e.}, the densities or colors are not constant over different time steps. By separating the static and dynamic voxels, the redundant computations caused by using a relatively heavy time-varying model to process the static components will be avoided, which accelerates the training and rendering.

A simpler solution for accelerate training is to separate the static and dynamic regions in pixel-level, i.e., using the temporal variance of pixels to produce static and dynamic ones. However this scheme is actually not feasible because we can only separate dynamic and static regions in training views using the ground truth. For rendering novel views, we can not get the pixel variance for rendering since we have no ground truth in novel views. Thus a feasible solution is to learn the voxel-level temporal variance which is shared for all possible views. In addition, separating in the voxel-level is more efficient compared to pixel-level, even if we have an oracle to make the pixel-level separation feasible in novel views. This is because not all voxels projected to a dynamic pixel are dynamic, there will be only a small fraction of voxels around the object surfaces are actually dynamic. Therefore, the voxel level separation will produce much fewer dynamic queries. 

\begin{figure}[t]
\centering
\includegraphics[width=.9\linewidth]{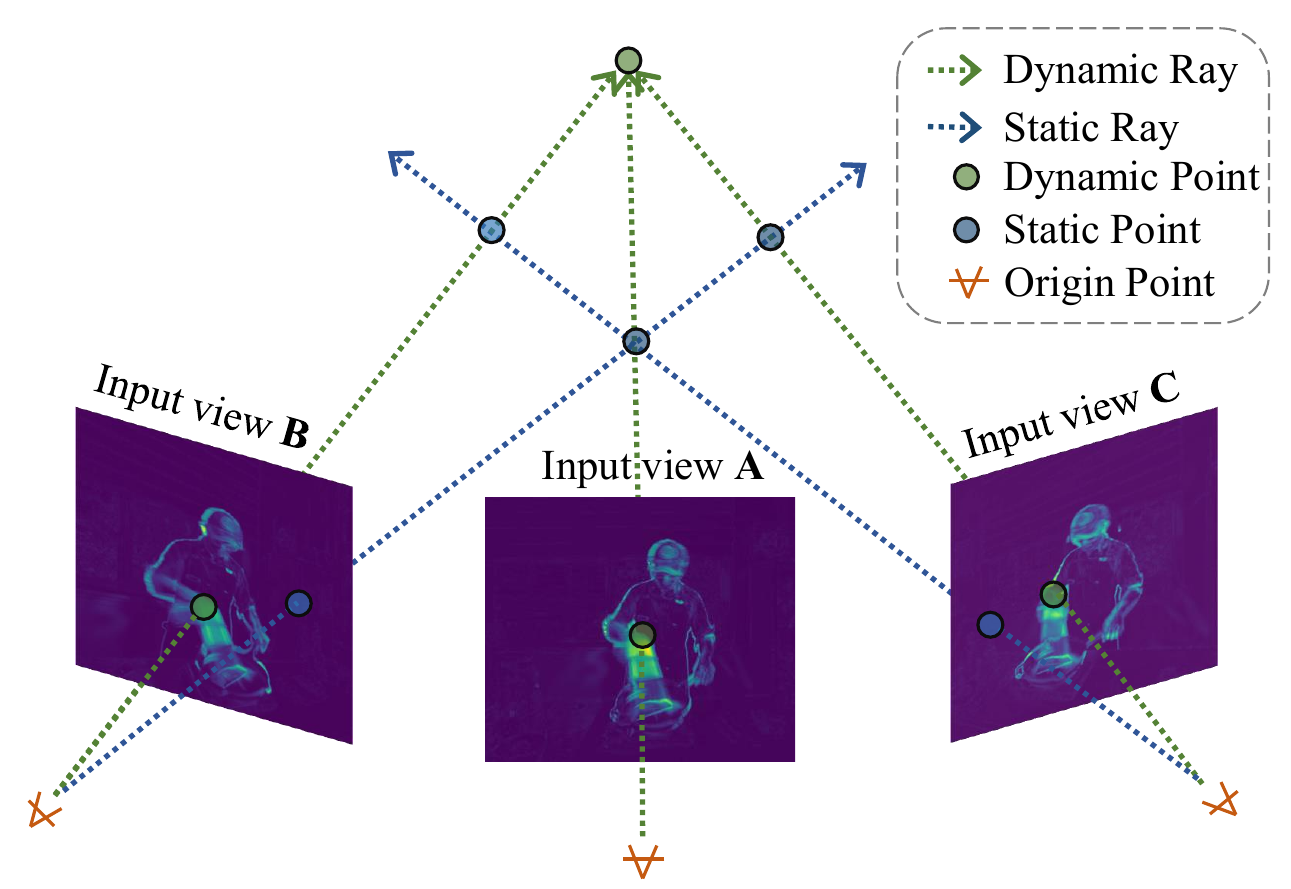}\\
\vspace{-.5em}
\caption{Implicit interaction of multiple rays to decide which point is dynamic. For a static ray (\textcolor[RGB]{47,85,151}{blue line}), all points are set to be low dynamic. For a dynamic ray (\textcolor[RGB]{84,130,53}{green line}), at least one point is dynamic. The intersection of multiple dynamic ray is more likely to be a dynamic point, which is also physically intuitive.}
\label{fig: variation}
\vspace{-1.2em}
\end{figure}

To perform the voxel-level separation, we utilize the pixel-level temporal variances from training videos as the supervision to estimate the voxel-level variances. The pixel-level (or ray-level) temporal variances of different videos are shown in \cref{fig: variation}. Formally, given a ray $\bm{r}(\mathrm{s}) = \bm{o} + \mathrm{s} \cdot \bm{d}$ with origin $\bm{o}$ and direction $\bm{d}$, the corresponding pixel color at time step $t$ is defined as $C(\bm{r}, t)$. Then the pixel-level temporal variance $D^2(r)$ is formalized as
\begin{small}
$$
    D^2(\bm{r}) = \frac{1}{T}\sum_{t=1}^{T} (C(\bm{r}, t) - \bar{C}(\bm{r}))^2, \quad
    \bar{C}(\bm{r}) = \frac{1}{T}\sum_{t=1}^{T}C(\bm{r}, t),
$$
\end{small}
where $\bar{C}(\bm{r})$ is the mean color of pixel corresponding to the ray $\bm{r}$. 
For identifying the dynamic pixels, the standard deviation $D(\bm{r})$ is binarized to $M(\bm{r})$ with a threshold $\gamma$ to provide pixel-level dynamic supervision, \textit{i.e.} $M(\bm{r})=1$ if $D(\bm{r}) \geq \gamma$, else $M(\bm{r})=0$. In this way, we judge that a ray $\bm{r}$ is dynamic if $M(\bm{r})=1$. Next, we use the $M(\bm{r})$ as supervision to estimate the voxel-level variations $\mathcal{V}$.

The relations between the pixel-level variance and voxel-level variance lie in the following aspects: (1) If a pixel is static, then all voxels passed through by the ray corresponding to the pixel should be static in most cases (We will discuss some special cases which violate this rule later in this subsection.). (2) If a pixel is dynamic, then at least one of the voxels passed through by the corresponding ray is dynamic. \cref{fig: variation} shows the two situations. With the above two relations, we design the variation field, which is denoted as $\mathcal{V} \in \mathbb{R}^{N_x \times N_y \times N_z}$ to represent the voxel-level temporal variance. Specifically, we uniformly sample $N_s$ points from the near plane to the far plane in $\bm{r}$, and build the following equation to satisfy the two relations mentioned above:
\begin{equation}
\small
    \hat{M}(\bm{r}) = \bm{s}( \mathrm{max}(\{\mathcal{V}_{\bm{r}(s_i)}| i \in \{1,...,N_s\}\})),
    \label{max}
\end{equation}
where $\hat{M}(\bm{r})$ is an estimation of $M(\bm{r})$, and $\bm{s}$ is the sigmoid function. Then we train the variation field by minimizing the following binary cross-entropy loss:
\begin{equation}
\footnotesize
    \mathcal{L}_{v} = \mathbb{E}_{\bm{r}}\left[
    - M(\bm{r}) {\rm log}(\hat{M}(\bm{r}))
     - (1-M(\bm{r})){\rm log}(1-\hat{M}(\bm{r}))\right].
\label{var_opt}
\end{equation}
By optimizing the above loss function to all rays, we can get the learned variation field $\mathcal{V}$. The training of the variation field is very efficient, usually taking \textit{less than 30 seconds}.

The maximization operation well formulates the relations between a pixel and its corresponding voxels. If a pixel is static, then the equations of \cref{max} and \cref{var_opt} will force all voxels passed through by the corresponding ray to be static ($\mathcal{V}_{x,y,z} = 0$). If a pixel is dynamic, \cref{max} requires at least one of the corresponding voxels (\textit{i.e.}, the max value of the voxel variances) to be dynamic ($\mathcal{V}_{x,y,z} = + \infty$). Although we provide no information about which specific voxels in a dynamic ray are dynamic, the implicit interaction of multiple different rays will force the solution to be physically reasonable. To explain this, we focus on the observable voxels which at least passed through by one ray. If a point $(x,y,z)$ is passed through by at least one static ray, then $\mathcal{V}_{x,y,z}$ will tend to be optimized to be close to zero. If a point $(x,y,z)$ is only passed through by dynamic rays, and not occluded by other dynamic voxels, then $\mathcal{V}_{x,y,z}$ will be optimized to $+ \infty$. This is because without occlusion, the above point of $(x,y,z)$ along the dynamic rays will be passed by other static rays (the front space along these rays are observable from other views). We illustrate this situation in \cref{fig: variation}.

\def\infmask{\dot{{\rm M}}}
\textbf{Inference.} After the training process, the temporal variation at a specific 3D position $(x,y,z)$ is $\mathcal{V}_{x,y,z}$, which is easily acquired by interpolating the discrete variation field. We then identify a voxel in the scene as dynamic if $\mathcal{V}_{x,y,z}$ is larger than a hyper-parameter $\beta$, and as static if it is smaller than $\beta$. Formally, we will get a dynamic mask $\infmask \in \{0,1\}^{N_x \times N_y \times N_z}$, which will be used to split sampling points in a ray into static points and dynamic points. 
We evaluate the effectiveness of this inference method in the test views and find that the recall and precision are reasonable for splitting the dynamic and static parts (recall: 0.97, precision: 0.94 when $\beta=0.9$). 
Although the recall seems sufficient to retrieve most dynamic parts, we empirically find some false negatives in the rendering images affect the rendering quality. To address this problem, we use a ray-wise max-pooling operation to identify the points near to a dynamic point as dynamic. The kernel size of max-pooling is set to $k_m=21$, and the stride is set to $1$. In this way, the recall is very closed to $1$. Although many hyper-parameters are incorporated, we have empirically found that the thresholds $\gamma$ and $\beta$ are not sensitive over a wide range of reasonable values.

\begin{figure}[t]
\centering
\includegraphics[width=1.0\linewidth]{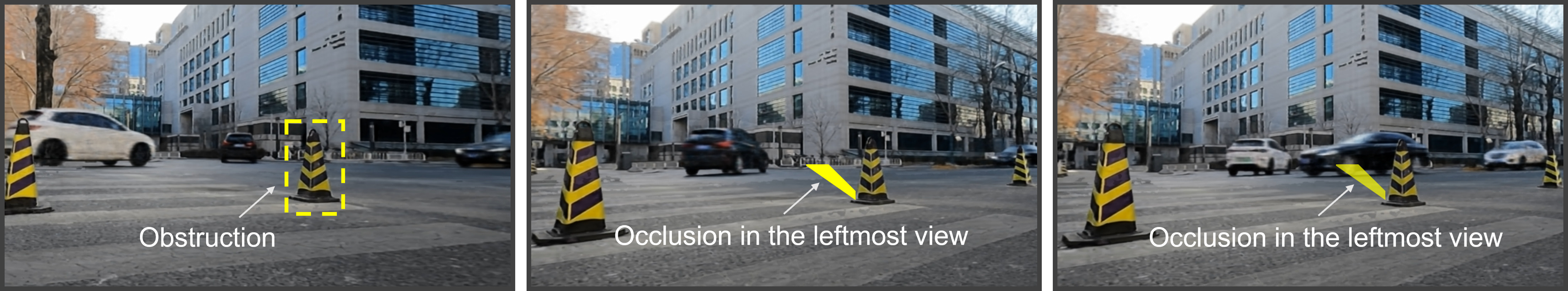}\\
\vspace{0.2em}
\caption{Impact of occlusions to the variation field.}
\label{fig: occlusion}
\vspace{-1.8em}
\end{figure}

\textbf{Discussion.} There may be some situations in which the rules (1) are broken due to occlusion. Specifically, when a dynamic voxel is occluded by some static voxels, the occluded parts should not be classified as static, which makes the 2D supervision noisy. However in practice, we found the learning-based separation has a certain degree of tolerance for this situation. We conducted experiments to verify this and present the results in \cref{fig: occlusion}. The occluded region is visualized in the leftmost view (the area behind the static roadblock). Although the dynamic region marked in yellow is occluded in the left view, it is actually classified as dynamic region. This can be inferred from another view where the occluded region has changed, as illustrated in the middle and right images in \cref{fig: occlusion}). The variation field is learning-based and will learn a solution that satisfies most constraints. If it forces some dynamic voxels occluded by static voxels to be $0$, then the loss function from other visible views will be high. As a result, the learning-based process tends to assign a ``middle solution" to voxels with inconsistent supervisions from different views. We also attemped to use the transmittance as a weight (in a way of volumetric rendering) to learn the variation field which can explicitly handle the occlusion problem. However, we found a large efficiency drop with similar performance. As a result, we use the proposed variation field and find this formulation works well for most scenes, including some challenging scenes with large areas of motions.

\subsection{Training of Mixed Neural Voxels}
With the help of the variation field, we can split a scene into dynamic voxels and static voxels. To reduce redundant computations, we use the lightweight static model described in \cref{sec: static_model} to compute the densities and colors for static voxels and the dynamic model described in \cref{sec: dy_model} to compute the densities and colors for dynamic voxels. The overall architecture is illustrated in \cref{fig:arch}.

Specifically for a given ray $\bm{r}(s) = \bm{o} + s\bm{d}$ with origin $\bm{o}$ and view direction $\bm{d}$, we apply stratified sampling from the near to the far planes and get $N_s$ points. 
Then the $N_s$ points are separated into static and dynamic ones by inferring these points with the proposed variation field. For the static points, we pass them into the static branch to retrieve the colors and densities. For the dynamic points, we pass them into the dynamic branch together with a deferred time query $\omega_t$ to retrieve the corresponding properties. After that, we merge the static points and dynamic points according to their order. Then we apply volumetric rendering to the merged points to obtain the rendered color, which is formulated as
\vspace{-0.4em}
\begin{equation}
\label{eq: render}
    C(\bm{r}, t) = \sum_{i=1}^{N_s} T_{i,t} \cdot (1-{\rm exp}(- \sigma_{i,t} \delta_i)) \cdot c_{i,t} ,
    \vspace{-0.2em}
\end{equation}
where $\mathrm{T}_{i,t}$ is the accumulated opacity (or transmittance):
$\mathrm{T}_{i,t} = {\rm exp}(-\sum_{j=1}^{i-1} \sigma_{j,t} \delta_j)$, and $\delta_i$ is the distance between adjacent samples. Given the ground truth color $C_{g}(\bm{r},t)$, an $l2$-loss is employed to train the model:
\begin{equation}
\vspace{-0.2cm}
    \mathcal{L} = \mathbb{E}_{(\bm{r},t)} \left[\Vert C_g(\bm{r},t) - C(\bm{r},t) \Vert_2^2 \right] .
\end{equation}

For both static and dynamic branches, we omit the computation of color for points whose densities are close to zero, which is a widely adopted pruning strategy \cite{liu2020neural, yu2021plenoxels, chen2022tensorf}.

\def\flops{FLOP$_{\rm sta}$\xspace}
\def\flopd{FLOP$_{\rm dyn}$\xspace}
\textbf{Efficiency analysis.} We define the proportion of dynamic points in a scene as $\lambda$ ($\approx 0.05$ for most scenes). Besides, the FLOPs of static and dynamic branches are denoted as \flops and \flopd, respectively. Then the total FLOPs of \ourmethod are \flops + $\lambda \cdot $\flopd. Empirically, \flopd$/$\flops ranges from $50$ to $100$ with different reasonable dimension settings. Then the acceleration ratio of splitting static and dynamic models is \flopd$/ $(\flops + $\lambda$ \flopd) $\approx 10$. In practice, the actual speedup with a 3090 GPU is about $5$, the inconsistency between analysis and experiment may come from the GPU features, which are friendly to a more consistent network.
\begin{figure*}[t]
\centering
\includegraphics[width=1.0\linewidth]{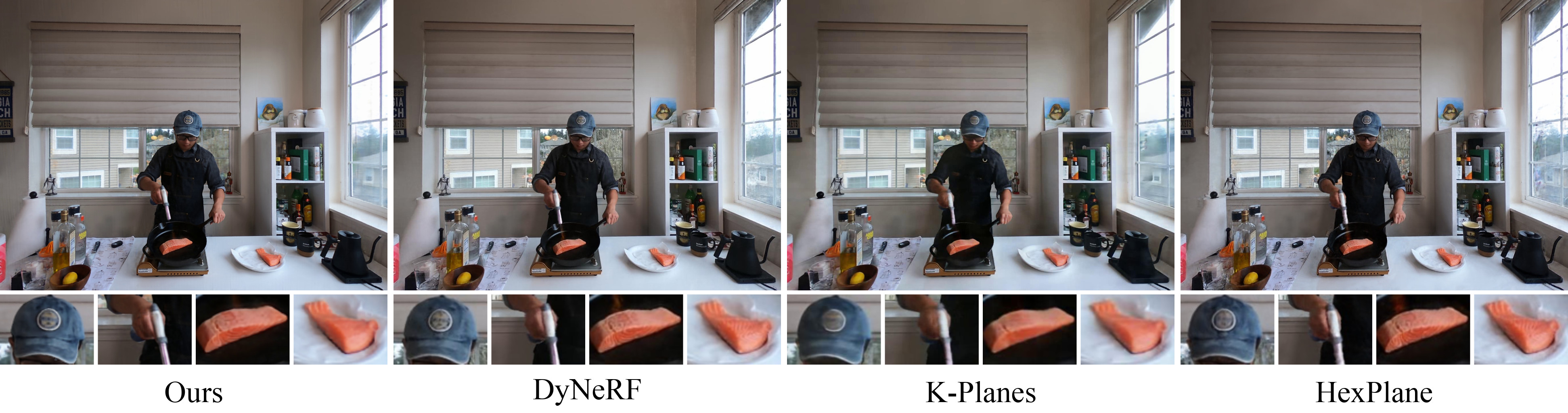}\\
\vspace{-.2em}
\caption{Visual comparisons with state-of-the-art methods. K-Planes \cite{fridovich2023k} and HexPlane \cite{cao2023hexplane} are concurrent works. We have selected four representative patches to better inspect the details. Our method performs well on reconstructing details and capturing movements.}
\label{fig:cmp_method}
\vspace{-1.2em}
\end{figure*}
\subsection{Implementation Detail}
The voxel-grid representations require large GPU memory to store the cubically growing voxel numbers. To implement the voxel-grid representations more memory efficient, we use the tensor factorization technique proposed in TensoRF \cite{chen2022tensorf} to reduce the memory footprint. In this way, a 3D tensor is factorized into the outer product of a vector and a 2D matrix. We factorize all the voxel-grid tensors, including static voxels, dynamic voxels and the variation field. With the help of tensor factorization, the learned model costs about 500MB for a 300-frame multi-view video scene. For the voxel resolutsions, we follow \cite{chen2022tensorf} to start from an initial low resolution of $256^3$, and upsample the resolution at steps 1500, 2000, 2500, and 2750 with a linear increase in the log space. The final resolution is set to $640^3$. Once the resolution is changed, we re-train the variation field, which only takes about 15-30s. The voxel-grid feature dimension is set to $27$, and the hidden state of MLP is set to $512$. For training, we use Adam \cite{kingma2014adam} optimizer with a learning rate of $0.02$ for voxels and $3e-3$ for MLPs. The total variation loss \cite{yu2021plenoxels} is incorporated as a regularization to encourage the space smoothness.
\section{Experiments}
\subsection{Experiment Setting} 
\textbf{Dataset.} We validate our method on two datasets: (1) The Plenoptic Video Dataset \cite{li2022neural}, which consists of 6 publicly accessible scenes: coffee-martini, flame-salmon, cook-spinach, cut-roasted-beef, flame-steak and sear-steak. We conduct experiments on all six scenes. Each scene contains 19 videos with different camera views. The dataset contains many challenging scenes including objects with topology changes, objects with volumetric effects, various lighting conditions, etc. (2) Our proposed dataset including two more complex dynamic scenes: moving-cars and solving-rubik. 
The moving-cars scene features several vehicles passing across the screen, with significant motion and displacement. Meanwhile, in the solving-rubik scene, a man solves a Rubik's cube at a rapid pace, averaging 4 rotations per second, providing an opportunity to evaluate the model's ability to capture swift movements. The collection procedures used are similar to those of DyNeRF. More details are presented in the appendix.

For training and evaluation, we follow the experiment setting in \cite{li2022neural} that employs 18 views for training and 1 view for evaluation. 
To quantitatively evaluate the rendering quality on novel views, we measure PSNR, DSSIM and LPIPS\cite{zhang2018unreasonable} on the test views. We also provide more metrics in the appendix including FLIP \cite{andersson2020flip} and JOD \cite{kiran2017towards}, which we find the comparisons are similar with PSNR and LPIPS. We follow the setting of \cite{li2022neural} to evaluate our model frame by frame. For videos consisting of equal or more than 300 frames, we evaluate our model every 10 frames \cite{li2022neural} to calculate the frame-by-frame metrics except for the JOD metrics, which requires a stack of continuous video.
\begin{table}[t]\vspace{0em}
\caption{Results on our collected dataset, including two scenes.}
\vspace{-0.5em}
\begin{center}
\tablestyle{4pt}{1.05}
\begin{tabular}{llccc}
\shline
Scene & Model & PSNR$\uparrow$ & DSSIM$\downarrow$ & LPIPS$\downarrow$ \\
\shline
\multirow{4}{*}{\textbf{Moving-Cars}} & \ourmethod-S & 18.72 & 0.251 & 0.689 \\
& \ourmethod-M & 18.97 & 0.228 & 0.552  \\
& \ourmethod-L & 18.89 & 0.222 & 0.540  \\
& \ourmethod-X & \textbf{19.11} & \textbf{0.210} & \textbf{0.516}  \\
\shline
\multirow{4}{*}{\textbf{Solving-Rubik}} & \ourmethod-S & 25.39 & 0.065 & 0.339 \\
& \ourmethod-M & 26.05 & 0.059 & 0.275 \\
& \ourmethod-L & 26.28 & 0.055 & 0.241  \\
& \ourmethod-X & \textbf{26.80} & \textbf{0.047} & \textbf{0.209}  \\
\shline
\end{tabular}
\end{center}
\vspace{-2.6em}
\label{tab: resour}
\end{table}

\textbf{Training Schedules.} 
For evaluating the effect of training time, we train \ourmethod with different configurations shown in \cref{tab: model}. 
The configurations vary in terms of training iterations and the number of sample points per ray. By default, the step size for sampling points is set to four times of the voxel width. The $8\times$ means that there will be eight times as many sampling points compared to the default.
\begin{table}[H]\vspace{-0.6em}
\caption{Different training configurations of \ourmethod.}
\vspace{0.3em}
\begin{center}
\tablestyle{3pt}{1.05}
\begin{tabular}{lccc}
\shline
Model  & Iterations & Sampling points & Training Time \\
\shline
\ourmethod-S & 5000  & 1 $\times$ & 15 min\\
\ourmethod-M & 12500 & 1 $\times$ & 40 min\\
\ourmethod-L & 25000 & 1 $\times$ & 80 min\\
\ourmethod-X & 50000 & 8 $\times$ & 300 min \\
\shline
\end{tabular}
\end{center}
\vspace{-1.6em}
\label{tab: model}
\end{table}
\subsection{Results}
\textbf{Quantitative results and comparisons.}
For quantitative results, we present the metrics and compare with other methods in \cref{tab: speed}. Compared with the previous state-of-the-art method DyNeRF, we reduce the training time from 1.3K GPU hours to 15 minutes, making the training of complex dynamic scenes more practical. For rendering, the \ourmethod has a 37.7 fps rendering speed for 1K resolution. Compared with concurrent works, \ourmethod requires less training time and achieves faster rendering speed, while achieving competitive PSNR and LPIPS. For example, with only 15 minutes of training, \ourmethod achieve 31.03 PSNR which is comparable to other methods trained for hours. With sufficient training, all metrics are further improved. 
For the quantitative results on our collected more complex scenes, we present them on \cref{tab: resour}. 

\begin{table}[t]
\caption{Quantitative results comparisons. All metrics are measured on 300-frame scenes. We also report the training time, rendering speed (FPS) and model size. $\ast$ Note DyNeRF is trained on 8 GPUs, while others are trained on one GPU.}
\vspace{-1.4em}
\begin{center}
\tablestyle{2.5pt}{1.05}
\begin{tabular}{lx{27}x{19}x{30}x{23}x{27}x{25}}
Method & Train & Render & Size & PSNR$\uparrow$ & DSSIM$\downarrow$ & LPIPS$\downarrow$\\
\shline
DyNeRF\cite{li2022neural} & 7 days$\ast$ & - & 28 MB & 29.58 & 0.0197 & 0.083\\
\shline
\multicolumn{6}{l}{Concurrent work}\\
StreamRF\cite{li2022streaming} & 75 min & 8.3 & 5310MB & 28.26 & - & - \\
NeRFPlayer\cite{song2022nerfplayer} & 360 min & 0.05 & - & 30.69 & 0.034 & 0.111 \\
Hyperreel\cite{attal2023hyperreel} & 540 min & 2.0 & 360 MB & 31.10 & 0.036 & 0.096 \\
K-Planes\cite{fridovich2023k} & 108 min & - & - & 31.63 & 0.018 & -\\
HexPlanes\cite{cao2023hexplane} & 720 min & - & 200MB & 31.71 & \textbf{0.014} & 0.075 \\
\shline
\ourmethod-S & 15 min & 37.7 & 500 MB & 31.03 & 0.022 & 0.129\\
\ourmethod-M & 40 min & 37.7 & 500 MB & 31.22 & 0.019 & 0.102 \\
\ourmethod-L & 80 min & 37.7 & 500 MB & 31.34 & 0.017 & 0.096 \\
\ourmethod-X & 300 min & 4.6 & 500 MB & \textbf{31.73} & 0.015 & \textbf{0.064} \\
\shline
\end{tabular}
\end{center}
\label{tab: speed}
\vspace{-2.5em}
\end{table}

\begin{figure}[t]
\centering
\includegraphics[width=1.0\linewidth]{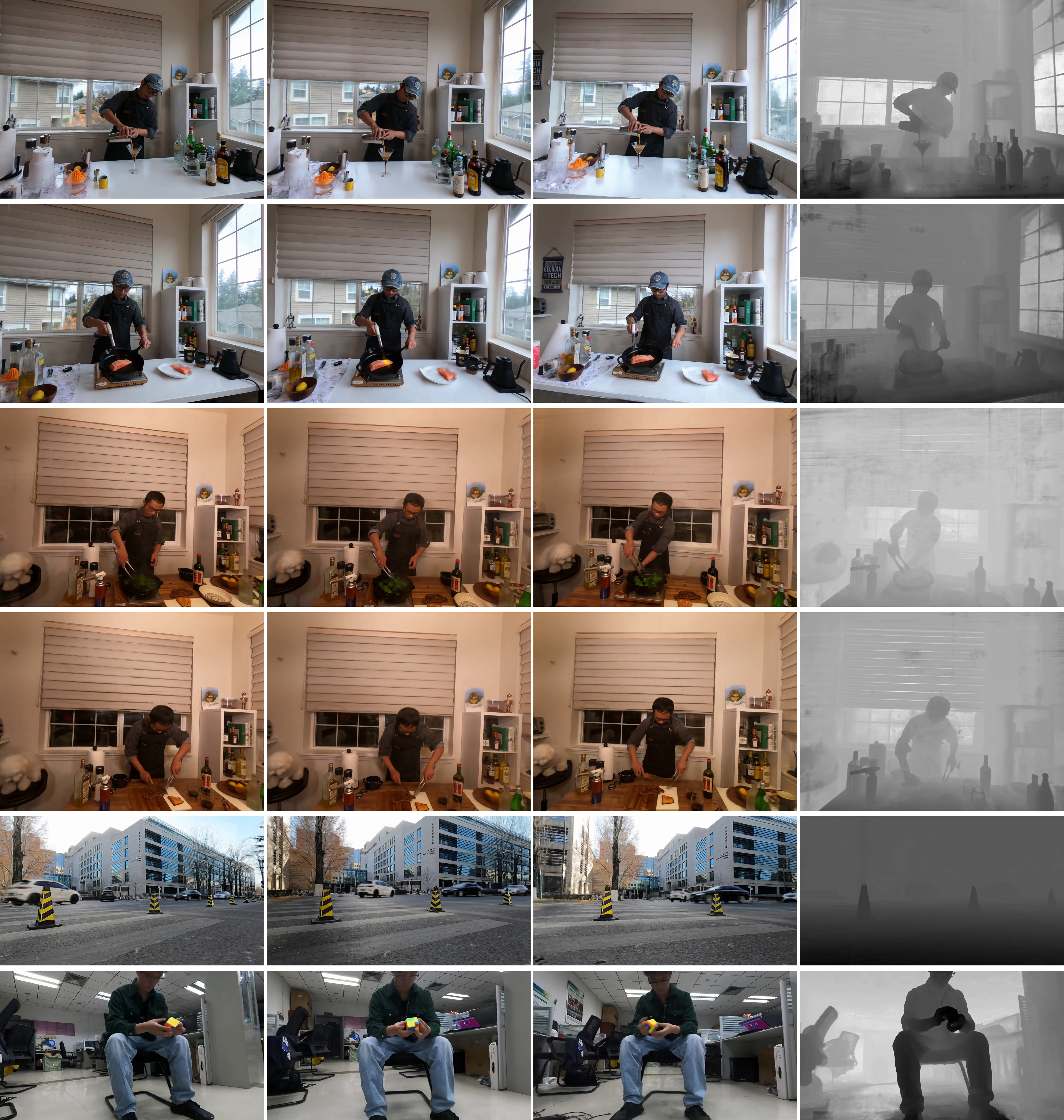}\\
\caption{Novel view synthesis of \ourmethod. We select some frames at different views. The last column demonstrates the normalized depth. We provide videos at the supplemental material.}
\label{fig:gallery}
\vspace{-1.6em}
\end{figure}

\textbf{Qualitative results and comparisons.}
\cref{fig:gallery} demonstrates the novel view rendering results on different dynamic scenes. The first four rows are novel view videos from Plenoptic Video Dataset \cite{li2022neural}. The last two rows present the novel view videos from our collected two more complex dynamic scenes. The results show that our method can achieve near photo-realistic rendering quality. We provide the video results at the supplementary material. For qualitative comparisons, we show them in \cref{fig:cmp_method}. \ourmethod can better reconstruct the moving object (the firing gun) and textual details like the hat and the salmon stripes. 

\begin{figure}[t]
\centering
\includegraphics[width=1.0\linewidth]{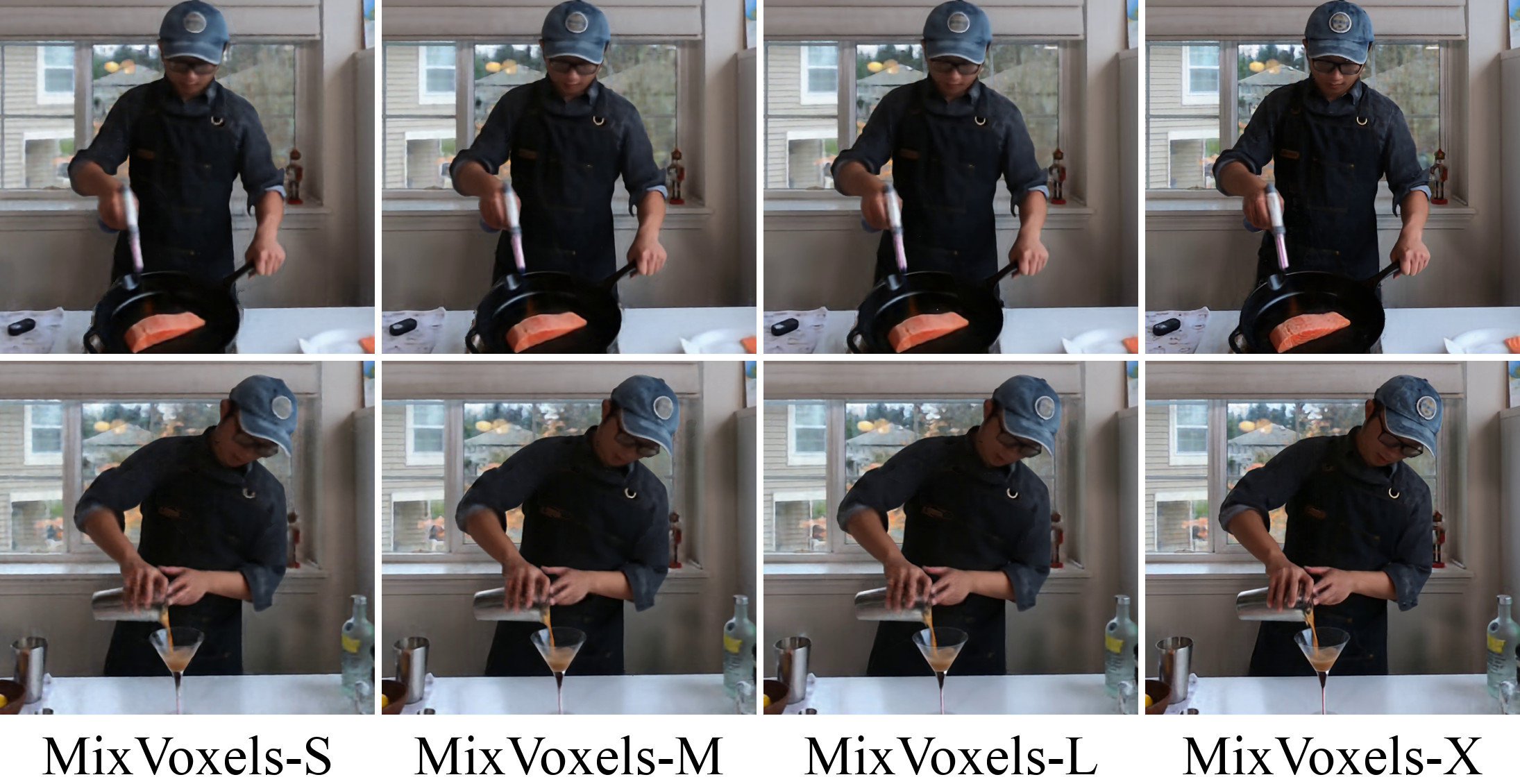}\\
\caption{Qualitative demonstration of different training schedules. Longer training helps better reconstruct the high-dynamic parts.}
\label{fig:schedule_quality}
\vspace{-0.5em}
\end{figure}

We further investigate the relations between rendering efficiency and rendering quality. As shown in the lower part of \cref{tab: speed}, it was observed that an increase in training time leads to improvements in both PSNR and LPIPS. Longer training facilitates the reconstruction of sharp boundaries and fine details. Visual comparisons presented in \cref{fig:schedule_quality} reveals that 15 minutes of training produces satisfactory recovery of most scene components but resulted in blurry motion details. With longer training, the moving objects become clearer with a distinct boundary.

\subsection{Ablation Study}
In this subsection, we empirically justify the design of \ourmethod by ablating or modifying several key features. We also provide analysis that intuitively explains the ablations. We conduct all experiments in this subsection on the coffee-martini scene, which we find is typical for demonstrating the fast-moving complex objects.

\begin{figure}[t]
\centering
\includegraphics[width=1.0\linewidth]{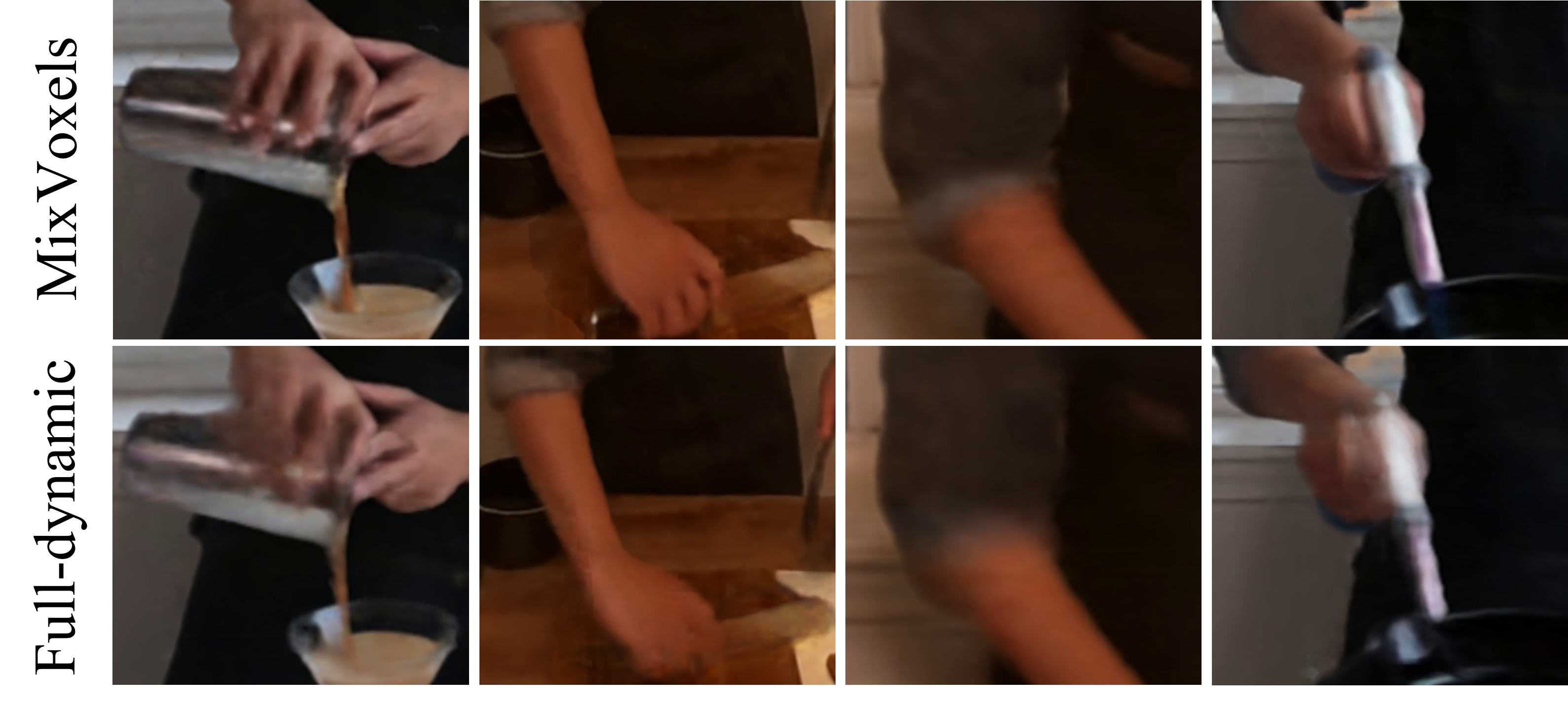}\\
\vspace{-0.2em}
\caption{Qualitative comparison of \ourmethod and the full-dynamic model. Training with full-dynamic models with the same iterations can not well reconstruct the motion details.}
\label{fig:ablation_split}
\vspace{-0.5em}
\end{figure}

\textbf{Ablation on splitting voxels.}
To study the effect of splitting static and dynamic voxels, we compare \ourmethod with a full-dynamic voxel-grid representation, where all points are processed by the dynamic model. \cref{tab: absplit} shows the comparisons. With the same training iterations, the full-dynamic model is more time-consuming, which is intuitive because it processes all voxels with the dynamic models. \cref{fig:ablation_split} shows the qualitative comparison. The full-dynamic model recovered blurred motions. We speculate the reason is because the large area of static regions affects the capturing of dynamic information. The network will be biased by most static voxels with no motions and tend to learn low-frequency information. 

\begin{table}[t]
\caption{Ablation on the mixed voxels. Training with the full-dynamic voxel model hurts both efficiency and efficacy.}
\vspace{-0.6em}
\begin{center}
\tablestyle{4pt}{1.05}
\begin{tabular}{lx{18}x{22}x{26}x{26}x{25}x{18}}
Method & Time  & PSNR$\uparrow$ & DSSIM$\downarrow$ & LPIPS$\downarrow$ & FLIP$\downarrow$ & JOD$\uparrow$ \\
\shline
Full-Dynamic & 2.5h & 28.36 & 0.036 & 0.2236  & 0.1196 & 7.44 \\
\ourmethod & \textbf{0.6h} & \textbf{29.47} & \textbf{0.026} & \textbf{0.1167} & \textbf{0.1223} & \textbf{7.99} \\
\shline
\end{tabular}
\end{center}
\vspace{-1.4em}
\label{tab: absplit}
\end{table}

\begin{table}[t]
\caption{Ablation on three different methods for time query. We only substitute the time query with different method, and train them on the proposed \ourmethod framework.}
\vspace{-0.6em}
\begin{center}
\tablestyle{3.6pt}{1.05}
\begin{tabular}{lx{22}x{22}x{26}x{26}x{25}x{18}}
Method & Time & PSNR$\uparrow$ & DSSIM$\downarrow$ & LPIPS$\downarrow$ & FLIP$\downarrow$ & JOD$\uparrow$ \\
\shline
Concat & 58m & 28.95 & 0.037 & 0.2146 & 0.1294 & 7.44\\
Fourier  & 43m    & 28.67 & \textbf{0.029} & 0.1824 & 0.1286 & 7.60\\
Inner product & 40m & \textbf{29.47} & \textbf{0.026} & \textbf{0.1167} & \textbf{0.1223} & \textbf{7.99} \\
\shline
\end{tabular}
\end{center}
\vspace{-2.5em}
\label{tab: ab_head}
\end{table}

\textbf{Ablation on time query.}
We compare our inner product time query method with other variants: \textbf{(1)} Concatenation which concatenates the temporal embedding with the voxel features to be processed by an MLP. \textbf{(2)} Fourier head proposed by \cite{wang2022fourier} which reconstructs the dynamics in frequency-domain. \cref{tab: ab_head} shows the performance comparison. The concatenation query method is both space- and time-consuming. Querying one time step requires forwarding the fused features through the whole MLP. Limited by the GPU memory, we can only query 50 time steps per-iteration with the concatenation way, which harms the performance on high-dynamic regions. The Fourier head processes the features to predict the magnitudes of different frequency components and the performance is competitive, while it requires an additional inverse discrete Fourier transform to recover the information in the temporal domain. Overall, the inner product query is the simplest and most efficient way for querying.

\begin{figure}[t]
\centering
\includegraphics[width=1.0\linewidth]{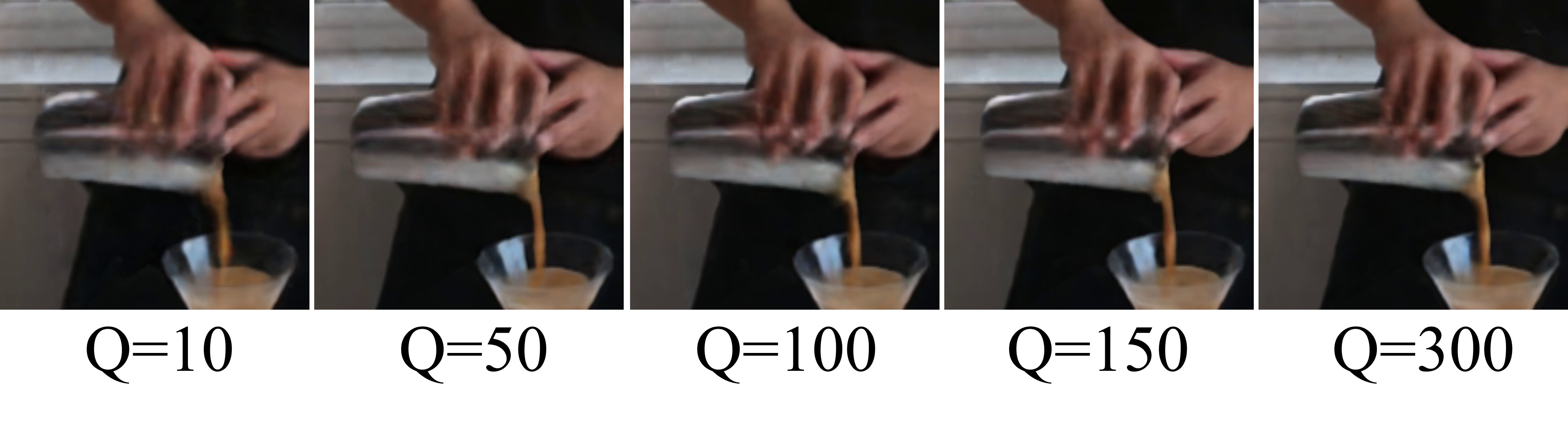}\\
\vspace{-0.5em}
\caption{Ablation on the number of time query denoted as Q.}
\label{fig:ablation2}
\vspace{-1.8em}
\end{figure}

\textbf{Number of time queries per-iteration.}
We empirically find that simultaneously querying multiple time steps in an iteration helps reconstruct the details of moving parts. \cref{fig:ablation2} demonstrates the effect of different numbers of time queries denoted as Q. With more time queries, the boundaries of the moving hand and the flowing coffee become clearer. Querying more time steps can provide dense supervision and make the model acquire global temporal information in every iteration, which accelerates the convergence speed. The effective inner product time queries allow adding more time queries with negligible increase in computation.

\subsection{Limitations}
\begin{figure}[H]
\centering
\vspace{-0.6em}
\includegraphics[width=1.0\linewidth]{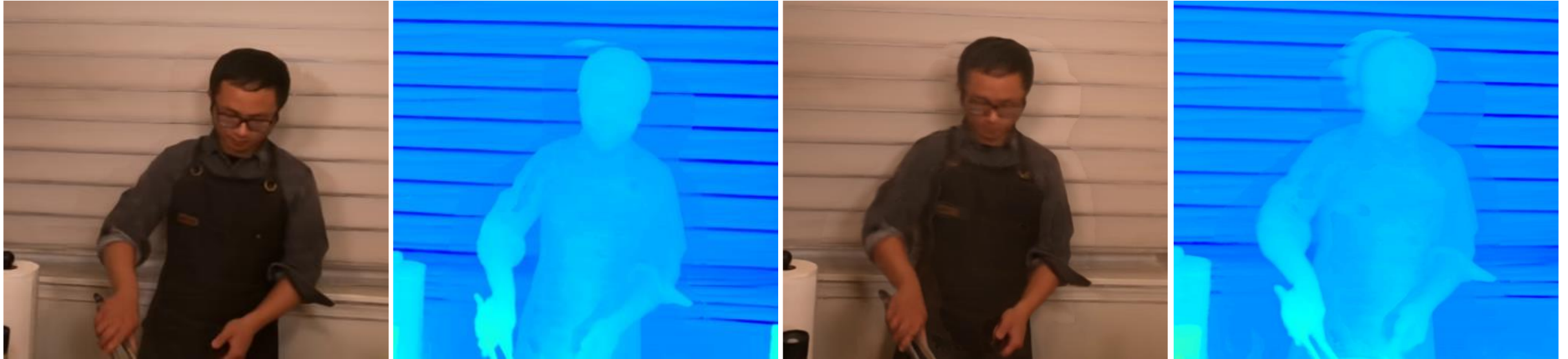}\\
\caption{Some inconsistent density and color predictions in the boundaries between dynamic and static regions.}
\label{fig:lim}
\vspace{-1em}
\end{figure}

Our method can synthesize novel view videos with a relative high quality. However, for some scenes with complex lighting conditions, some inconsistent property predictions may appear at the boundary between dynamic and static voxels, which is shown in \cref{fig:lim}. We suspect that the phenomenon is caused by the under-sampling of dynamic regions on scenes with some bad conditions. We will investigate ways to address the problem in future works.
\section{Conclusion}
This paper demonstrates a new method named \ourmethod to efficiently reconstruct the 4D dynamic scenes and synthesize novel view videos. The core of our method is to split the 3D space into static and dynamic components with the proposed variation field, and process them with different branches. The separation speeds up the training and makes the dynamic branch focus on the dynamic parts to improve the performance. We also design an efficient dynamic voxel-grid representation with an inner product time query. The proposed method achieves competitive results with only 15 minutes of training, making the training and rendering of complex dynamic scenes more practical. We believe the fast training speed will enable potentially useful applications that are bottlenecked by training efficiency.
\section{Acknowledgement}
This work was supported in part by the National Natural Science Fund for Distinguished Young Scholars under Grant 62025304.
{\small
\bibliographystyle{ieee_fullname}
\bibliography{mixvoxels}
}

\end{document}